\documentclass[conference]{IEEEtran}
\IEEEoverridecommandlockouts
\usepackage{cite}
\usepackage{amsmath,amssymb,amsfonts}
\usepackage{algorithmic}
\usepackage{graphicx}
\usepackage{textcomp}
\usepackage{xcolor}
\def\BibTeX{{\rm B\kern-.05em{\sc i\kern-.025em b}\kern-.08em
    T\kern-.1667em\lower.7ex\hbox{E}\kern-.125emX}}
\begin{document}

\title{Parallel Knowledge Transfer in Multi-Agent Reinforcement Learning\\
}

\author{\IEEEauthorblockN{1\textsuperscript{st} Yongyuan Liang}
\IEEEauthorblockA{\textit{Robotics Institute} \\
\textit{Carnegie Mellon University}\\
Pittsburgh, United States\\
liangyy58@mail2.sysu.edu.cn}
\and
\IEEEauthorblockN{2\textsuperscript{nd} Bangwei Li}
\IEEEauthorblockA{\textit{School of Mathematics} \\
\textit{Sun Yat-sen University}\\
Guangzhou, China \\
libw5@mail2.sysu.edu.cn}
}

\maketitle

\begin{abstract}
Multi-agent reinforcement learning is a standard framework for modeling multi-agent interactions applied in real-world scenarios. Inspired by experience sharing in human groups, learning knowledge parallel reusing between agents can potentially promote team learning performance, especially in multi-task environments. When all agents interact with the environment and learn simultaneously, how each independent agent selectively learns from other agents' behavior knowledge is a problem that we need to solve. This paper proposes a novel knowledge transfer framework in MARL, PAT (Parallel Attentional Transfer). We design two acting modes in PAT, student mode and self-learning mode. Each agent in our approach trains a decentralized student actor-critic to determine its acting mode at each time step. When agents are unfamiliar with the environment, the shared attention mechanism in student mode effectively selects learning knowledge from other agents to decide agents' actions. PAT outperforms state-of-the-art empirical evaluation results against the prior advising approaches. Our approach not only significantly improves team learning rate and global performance, but also is flexible and transferable to be applied in various multi-agent systems.
\end{abstract}


\section{Introduction}
Knowledge transfer is a common method in the general learning process of a new task or in a new environment. The educational behavior in human society is an advanced form of knowledge transfer. In a multi-agent system, when an agent in an unfamiliar environment and learn to get more reward, the knowledge from other experienced agents is beneficial to the agent. Reinforcement Learning (RL) \cite{kaelbling1996reinforcement}, as a popular framework, has been employed in sequential decision-making problems. And Transfer Learning (TL) \cite{taylor2009transfer} aims to improve learning through the learning experience from a related task, which also means knowledge reusing. In Reinforcement Learning domains, the source of informative knowledge varies from experienced agents(experts) to human guidance. In this paper, we work on applying knowledge transfer method in agents' behavior transfer for multi-agent team.

Cooperative Multi-agent Reinforcement Learning (MARL) has been applied in a series of meaningful problems such as multi-robot control \cite{matignon2012coordinated} and team-game playing \cite{le2017coordinated}. In Cooperative MARL, if an individual agent's learning is simply seen as independent RL with partial observations, the interactions between agents and the non-stationary environment will cause significant difficulties. To accelerate the team-wide learning efficiency and maximize the advantage of knowledge-transfer in the multi-agent domain, this paper targets the problem of optimizing knowledge-transfer between agents in Cooperative MARL under local constraints with a joint task or multiple tasks.

Our work is different from prior works that study inter-agent communication mechanisms in cooperative MARL \cite{lowe2017multi,sukhbaatar2016learning,foerster2016learning}. These works require a centralized critic and decentralized execution framework. Considering the scale of an agent team in the real scenarios, centralized training is challenging considering training stability and high computation complexity with large computational costs. Jiang \& Lu \cite{jiang2018learning} and Das et al. \cite{das2018tarmac} both proposed attention based communication protocol to exchange messages in MARL domains \cite{jiang2018learning,das2018tarmac}, but these approaches did not consider agents' behavior knowledge. Communication methods in MARL domains are always designed for solve the problem of efficient sharing partial observation information. In a fully decentralized system, we try to find a method to reuse agents' experience information to guide the unfamiliar agents with less communication costs for the goal of improving team-wide learning. Our work concerns behavior knowledge reusing problem between cooperative agents with a teacher-student framework in order to improve team-wide performance in learning process, relevant to multi-agent teaching. 

Regarding privacy constraints and communication cost in a multi-agent team, the main issues to be concerned are knowledge transfer decision, knowledge selection, and knowledge utilization. Invalid or confusing messages from other agents may cause a negative impact on agents' individual learning. Also, in an environment where information interaction is frequent, there is a danger of risk contagion, resulting in poor performance of the entire team. For instance, the phenomenon of "over-advising" mentioned in previous studies, has increased team-wide learning instability, especially in a team with more than two agents. In order to improve knowledge communication efficiency, we introduce an attention mechanism to dynamically distill knowledge from other agents' experience. Crucially, attention mechanism as a teacher selector is used to determine teacher agents' familiarity with the environment and current policy effectiveness for the student agent. Hence, The student agent have the flexibility to accept the optimal advice from appropriate teachers for each time point. 

In this paper, we propose a more robust and reliable parallel knowledge-transfer model with high efficiency in MARL framework. We here state the following based settings in our work:
\begin{itemize}
    \item All agents simultaneously learn in an environment and make decisions with interactions with the environment and other agents.
    \item There is no optimal expert (good-enough agent) in the multi-agent team in the initial state.
    \item (In Parallel) For all agents, their local role of student or teacher is not fixed. An agent can use knowledge from other agents as a student and provide its own behavior knowledge as a teacher for students.
    \item All agents are learning for their local reward. But agents in the team are friendly to share knowledge. Our goal is to maximize the team-wide reward.
\end{itemize}

Our empirical results across a range of tasks and environments demonstrate the efficacy of our knowledge transfer architecture in multi-agent systems. We show that our attention selector is capable of teaching the student agent with the most confident advice from teacher agents. Compared with other multi-agent knowledge transfer frameworks, our approach successfully give rise to an obvious improvement in global performance and stability.

\section{Related Work}
As a long-standing topic in the field of Reinforcement Learning, Multi-agent Reinforcement Learning (MARL) \cite{bucsoniu2010multi} track has a series of works in various ways to improve the performance and efficiency of team coordination. Deep Reinforcement Learning \cite{mnih2015human} uses deep neural networks to approximate the policy and value functions of agents in the environment to address the problem of large-scale action-value space in RL. And Knowledge transfer method has been studied in several related fields including imitation learning, learning from demonstration \cite{le2017coordinated}, and inverse reinforcement learning \cite{hadfield2016cooperative}. Several works \cite{da2019survey} extended the source-target framework in transfer learning on reinforcement learning task. In the extensive teacher-student framework for transfer from expert policy to student policy, the student agent takes actions from the expert agent's advice. Student-initiated approaches mainly concern with the student’s decision value, such as Ask Uncertain \cite{clouse1997integrating} and Ask Important \cite{amir2016interactive}. In teacher-initiated approaches, teachers decide when to teach based on the comparison between student’s and teacher’s
learning experience, such as Importance Advising \cite{torrey2013teaching}, Early Correcting \cite{amir2016interactive}, and Correct Important \cite{torrey2013teaching}. Q-teaching \cite{fachantidis2018learning} designs teaching rewards to help teachers determine when to advise. Moreover, episode-sharing mechanism \cite{tan1993multi} helps agents share individual successful episodes to accelerate learning.

However, all of the above works require an expert (all-knowing teacher) as the best agent to guide the learning of other agents. Zhan et al. \cite{zhan2016theoretically} analyzed the case of negative transfer with the existence of a sub-optimal expert and present some theoretical results.

Recent works provide some solutions for multi-agent parallel advising problem:\\
\textbf{AdHocVisit and AdHocTD} \cite{da2017simultaneously} is an advisor-advisee framework without an expert in the multi-agent environment for agents learning simultaneously. The learning agents ask for advice and provide advising policy for other agents. Advisees use state visit counts to decide when to request advice and advisors evaluate their advice’ reliability through confidence metrics to decide when and what to provide for advisees. For advice selection, AdHocVisit and AdHocTD follow majority vote \cite{zhan2016theoretically}.\\
\textbf{LeCTR} \cite{omidshafiei2019learning} is a new teacher-student framework, which targets peer-to-peer teaching in order to solve advising-level problem. Each agent in system learns when
and what to advise. In LeCTR, teacher-student (advising-level) policies are trained using the multi-agent actor-critic approach (MADDPG) \cite{lowe2017multi}. LeCTR sets the advising-level policies as decentralized actor and uses a centralized action-value function as critic with advising-level reward. It is worth mentioning that LeCTR considers the communication cost in information exchange. LeCTR only works in two-player games.

Motivated by attention \cite{vaswani2017attention} introduced for information extraction in deep learning, attention mechanism has recently emerged in reinforcement learning framework \cite{oh2016control, choi2017multi}. In distributed MARL, Jiang \& Lu \cite{jiang2018learning} proposed an attentional communication method with independent actor-critic. Attention in this work encodes agents' individual observation before passing centralized communication channel. With centralized value estimate, TarMAC \cite{das2018tarmac} is a targeted communication architecture to generate agents' internal state representations as input of centralized critic. Iqbal \& Sha \cite{iqbal2018actor} introduced an attention-based  critic to select agents in centralized training. 

Although attention plays a core role in our idea, our motivation is different from aforementioned approaches. Our algorithm aims to reuse agents' accumulated knowledge in learning process but not limited to process individual observation with agents' interaction in multi-agent domains. Our approach is more effective and efficient to maximum team cooperation utility and flexible to extend in complex environments.

\section{Preliminaries}
In this work, we consider a decentralized multi-agent reinforcement learning scenario, multiple agents in cooperative team $\mathcal{G}$ simultaneously learn a joint task or multiple tasks. Our settings are formalized as a Decentralized POMDP (Dec-POMDP) in cooperative multi-agent system. All the agents in the environment receive local observation $o_t^i$ at each time step, and interact with the environment by executing local action. Agents then update their policy parameters according to the feedbacks (reward) given by the environment.

The system is described as $(\mathcal { I }, S , A , T , R , \Omega , O , \gamma)$,
\begin{itemize}
\item $S$ is a set of states,
\item $A$ is a set of joint actions, $\mathcal { A } = \times _ { i } \mathcal { A } ^ { i }$,
\item $T$ is a set of conditional transition probabilities $T(s ^ { \prime } | s, a)$ between states,
\item $R$: $S\times A\to \mathbb {R} $ is the global reward function.
\item $\Omega$ is a set of joint observations, $\boldsymbol { o } = \left\langle o ^ { 1 } , \ldots , o ^ { n } \right\rangle$
\item $O$ is a set of conditional joint observation probability, $P( \boldsymbol { o } | s ^ { \prime } , \boldsymbol { a } ) = \mathcal { O } \left( \boldsymbol { o } , s ^ { \prime } , \boldsymbol { a } \right)$,
\item $\gamma \in [0,1]$ is the discount factor.
\end{itemize}

At each time period, the environment is in some state $s\in S$. Agents take a joint action $\mathcal { A } \in A$. Then each agent receives a local reward $r^i_ { t } = \mathcal { R }^i \left( s _ { t } , \boldsymbol { a }^i _ { t } \right)$. The process repeats. 

The goal is for all agents to take actions at each time step that maximize the global expected future discounted reward: $E \left[ \sum _ { t = 0 } ^ { \infty } \gamma ^ { t } r _ { t } \right]$.

\subsubsection{Reinforcement Learning}\cite{kaelbling1996reinforcement} is a standard framework to achieve the above goal of MDP (or POMDP). The process of value based reinforcement learning is to learn a policy which can maximize agent’s final reward. Through collecting experience from environment, agent updates its value function$v^{\pi}(s)=\mathbb{E}_{\pi}\left[R_{t} | s_{t}=s\right]$and action value function$Q^{\pi}(s, a)=\mathbb{E}_{\pi}\left[R_{t} | s_{t}=s, a_{t}=a\right]$.

\subsubsection{Deep Q Learning (DQN)}\cite{mnih2015human} is a value-based Reinforcement Learning approach combined with deep neural networks, which learns the action value function (Q-value) in continuous environment using value function approximation. Q-Network updates by minimizing the loss:$L(\theta)=\mathbb{E}[(r+\gamma \max _{a^{\prime}} Q(s^{\prime}, a^{\prime} ; \theta)-Q(s, a;\theta)]$, and outputs the expected action value $Q(s, a;\theta)$.

\subsubsection{Deterministic Policy Gradient (DPG)}\cite{silver2014deterministic} is a policy-based Reinforcement Learning approach as an extension of policy gradient (PG) \cite{sutton2000policy}, which optimize policy by update policy parameters $\theta$ along
the gradient direction,
$$\nabla_{\theta} J(\theta), \nabla_{\theta} J(\theta)=\mathbb{E}_{s \sim p^{\pi}, a \sim \pi_{\theta}}[\nabla_{\theta} \log \pi_{\theta}(a | s) Q^{\pi}(s, a ; \theta)]$$. 
Then, DPG is extended to \textbf{Deep Deterministic Policy Gradient (DDPG)}\cite{lillicrap2015continuous}, with $\nabla_{\theta}J(\theta)=\mathbb{E}_{s \sim \mathcal{D}}[.\nabla_{\theta} \mu_{\theta}(a | s) \nabla_{a} Q(s, a ; \mu)|_{a=\mu_{\theta}(s)}]$ using deterministic policy when the variance of action probability distribution approaches 0.

\subsubsection{Attention Mechanism}is one of the most influential models which can be broadly interpreted as a vector of importance weights. It has recently been applied in reinforcement learning domains.

\section{Overview}
\begin{center}
\begin{figure}[t]
    \includegraphics[width=0.50\textwidth]{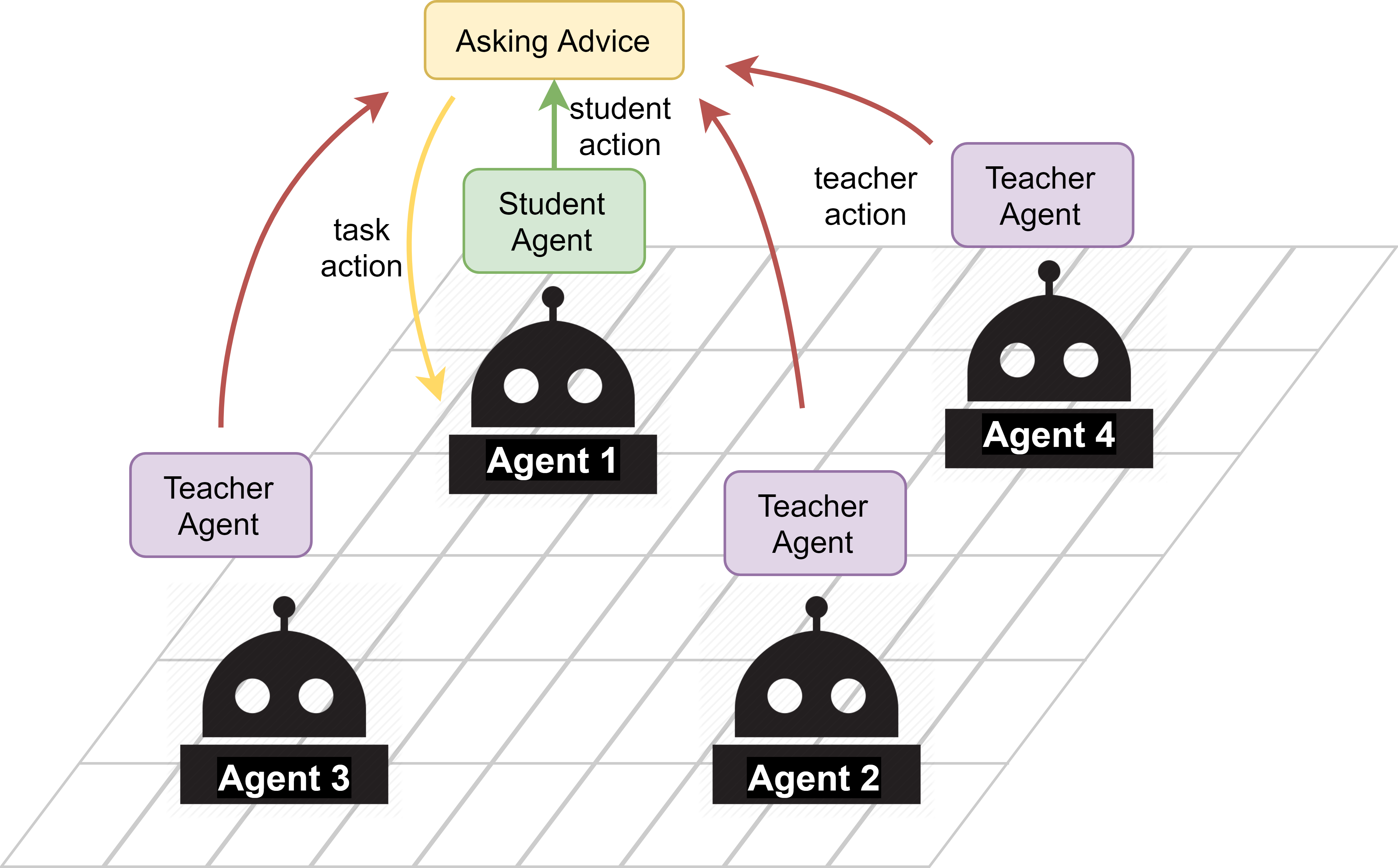}
    \caption{Overview}
    \label{fig:Overview}
\end{figure}
\end{center}

Our work targets knowledge reusing between agents in cooperative MARL, where all the agents in the environment are not good enough. In this section, we provide a story-level overview of our main idea. The overview of our motivating scenario is presented in Fig.~\ref{fig:Overview}.

Considering our settings, all the agents act in the environment and update their self policy parameters with local rewards from the environment. The actions executed by the agents is dictated by their self policy parameters. Now, we explore a novel knowledge transfer framework, PAT. In our framework, each agent has two acting modes, student mode and self-learning mode. Before each agent takes action, agents' student actor-critic modules decide agents' acting modes with agents' hidden states. It is not an ad-hoc design for specific domain. Agents should learn to ideally learn from other agents. 

In self-learning mode, agents take action based on their independent learned behavioral knowledge, which is represented as agents' behavior policy parameters. In this mode, agents' actions are independent of other agents' behavioral knowledge. All agents in self-learning mode are trained in an individual end-to-end manner using Deep Deterministic Policy Gradient \cite{lillicrap2015continuous} algorithm with an actor network and a critic network.

In student mode, if there are more than two agents in the system, a student agent receives multiple advice from other agents. We now refer to a new problem, teacher selecting, because not all teachers' knowledge is useful for the student agent. However, existing frameworks try to dodge this problem and have key limitation in the scenario where the number of agents is large, which also causes difficult in model transfer. We apply a soft attention mechanism in our work to select teachers' knowledge. The Attention Teacher Selector solves the problem by selecting contextual information in teachers' learning information and computing weights of teachers' knowledge. Considering from a different angle, our attentional module selectively transform the learning information from teachers with the target of solving student's problem. The attentional selecting approach is effective both in multi-task scenarios and joint task scenarios.

We make a few assumptions on agents' identities to support our framework. When an agent chooses student mode in $t$ time step, other agents in the environment automatically become its teachers and provide their behavior policy and learning knowledge to the student agent.

Our parallel setting means that An agent in student mode can be a teacher of the other agents. When agent $i$ is unfamiliar with its observation $m^i_t$, but at the same time, agent $i$ may be familiar with agent $j$'s observation $m^i_t$ because of $i$'s past trajectory, which means that agent i can be agent j's teacher. At this time step, agent $i$ is in student mode but it also is a teacher to transfer its knowledge to other agent. The core idea is agents' different learning experience. An student agent may have confidence in the other states and its behavior knowledge can help the other student agents. Our teacher selector module is designed to determine the appropriate teachers and transform teachers' local behavior knowledge into student's advising action. Moreover, our attention mechanism quantifies the reliability of teachers, so our scenarios do not need good-enough agents (experts).

In a high-level summary, because of agents' different learning experience, agents in a cooperative team are good at different tasks or different parts of a joint task. Knowledge Transfer is a framework to help agents solve unfamiliar task with experienced agents' learning knowledge. 

PAT's training and architecture details are presented in the next section.

\section{Attention based Knowledge-Transfer Architecture}
This section introduces our knowledge transfer approach with more design details of the whole structure and all training protocols in our framework. In our framework, each agent has two actor-critic model and an attention mechanism to support two acting modes. 

\subsection{Acting Mode}
Different from original individual agent learning, after receiving observation from the environment, agents in our framework need to choose their action mode before taking action.

At $t$ time step, agent $i$ reprocesses the observation from environment with a hidden LSTM (or RNN) unit, which integrates information (observation) from $i$'s observation history. The LSTM unit $l^i$ outputs agent's observation encoding $m^i_t$, which represents agent's hidden state. Here, $k$ is a scaling variable, which represents the time period covered in the hidden state. We will adjust $k$ depending on different types of games.
\begin{equation}
l^i : (o^i_{t-k}, a^i_{t-k}, ..., o^i_t) \rightarrow m^i_t
\end{equation}
Next, based on this step's memorized observation, $m^i_t$, agent $i$'s student actor network takes this step's memorized observation, $m^i_t$, as input and output agent $i$'s acting mode. Considering the efficiency of information exchange and communication cost, student actor is used to deciding agent $i$' confidence in $t$ time step. If $i$ has enough confidence with $m^i_t$, student actor chooses self-learning mode. Conversely, student actor chooses student mode and sends advice request to other agents. 

Student actor and student critic is a deep deterministic policy gradient model. The student actor network outputs the probability of choosing student mode. When the probability exceeds a threshold value, agent will choose student mode as a deterministic action. The threshold value is a variable depending on different types of games.

Student actor and student critic represent the acting mode choosing model which determines whether agent $i$ become a student and ask teacher agents for advice. We train student actor-critic using  a student reward ${\widetilde{r}}^i_t$.
\begin{equation}
\widetilde{r}_{t}^{i}=V\left(m^i_t; \theta'^{i}_{t}\right)-V\left(m^i_t; \theta^i_{t}\right)
\end{equation}
$\theta'^{i}_{t}$ and $\theta^i_{t}$, are agent $i$ policy parameters in student mode and self-learning mode. The student reward measures gain in agent learning performance from student mode. The sharing of student actor-critic network parameters allows this module learning effectively in environment and easily extending to other settings.

In our experiments, student actor-critic is trained with the trained Attention Teacher Selector. Agent $i$'s student critic is updated to minimize the student loss function: 

$\mathcal{\widetilde{R}}$ is agent $i$'s student policy transition set.
\begin{equation}
\begin{split}
    \mathcal{L}\left(\theta^{\tilde{Q}}\right)=\mathbb{E}_{m_t, w_t, \widetilde{r}_{t}, m_{t+1}\sim\mathcal{\widetilde{R}}}\left[\left(\tilde{y}_t-\widetilde{Q}\left(m_t, w_t | \theta^{\tilde{Q}}\right)\right)^{2}\right],\\
    \tilde{y}_t=\tilde{r}_t+.\gamma \widetilde{Q}(m_{t+1}, w^{\prime} | \theta^{\widetilde{Q^{\prime}}})|_{w^{\prime}=\widetilde{\mu}^{\prime}\left(m_{t+1} | \theta^{w^{\prime}}\right)}
\end{split}
\end{equation}
Student policy network is updated by ascent with the following gradient:

\begin{equation}
    {\nabla}_{\theta^{\widetilde{\mu}}}J=\mathbb{E}_{m_t, w_t \sim\mathcal{\widetilde{R}}}\left[\nabla_{w} \widetilde{Q}\left(m_t, w_t | {\theta}^{\widetilde{Q}}\right)|_{w_t=\widetilde{\mu}(m_t)} {\nabla}_{\theta^{\widetilde{\mu}}} \widetilde{\mu}\left(m_t | {\theta}^{\widetilde{\mu}}\right)\right]
\end{equation}

Here, $\widetilde{\mu}$ is agent $i$'s student policy, which is parameterized by $\widetilde{\theta}$

\begin{center}
\begin{figure}[t]
    \includegraphics[width=0.50\textwidth]{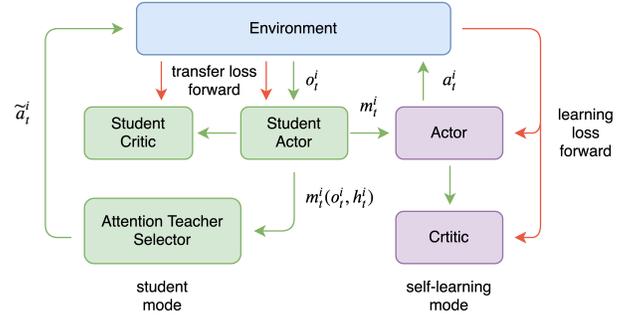}
    \caption{PAT Architecture}
    \label{fig:Architecture}
\end{figure}
\end{center}

\subsection{Student Mode}
\subsubsection{Attention Teacher Selector}

\begin{center}
\begin{figure}[t]
    \includegraphics[width=0.50\textwidth]{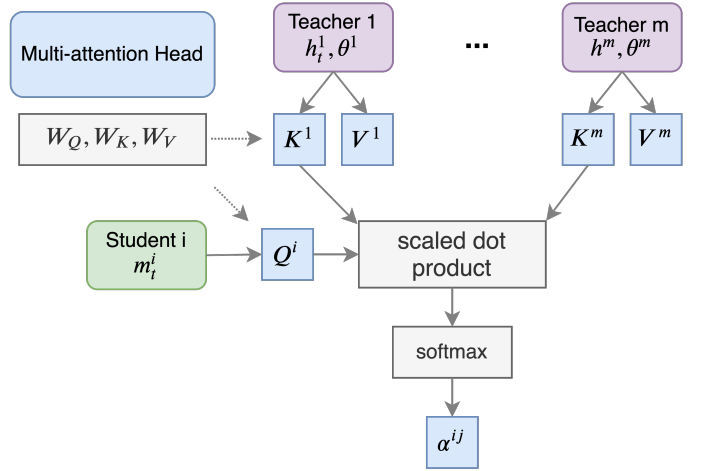}
    \caption{Attention based Knowledge Selection}
    \label{fig:Attention}
\end{figure}
\end{center}

Inspired by the similarity between source task and target task in transfer learning, we use attention mechanism to evaluate the task similarity between student and teachers and teachers' confidence of student's state. Therefore, each agent's  Attention Teacher Selector in student mode is used to select advice from teachers based on their similarity and confidence. The main idea behind our knowledge transfer approach is to learn the student mode by selectively paying attention to policy advice from other agents in the cooperative team. Fig.~\ref{fig:Attention} illustrates the main components of our attention mechanism.

We now describe the Attention Teacher Selector mechanism in agent student mode. The Attention Teacher Selector (ATS) is a soft attention mechanism as a differentiable query-key-value model \cite{graves2014neural,oh2016control}. After the student actor of student agent $i \in \mathcal{G}$ compute the memorized observation at $t$ time step and choose student mode, ATS receives the encoding hidden state $m^i_t$. Then, from other agents in the team as teacher agents, ATS receives the teachers' encoding learning history $h^j_t = l^j(o^j_1, a^j_1, ..., o^j_t)$ and encoding policy parameter $\theta^j$. 

Now, ATS computes a query $Q^i_t = W_Q m^i_t$ as student query vector, a key $K^j_t = W_K h^j_t$ as teacher key vector, and a value $V^j_t = W_V \theta^j$ as teacher policy value vector, where $W_K, W_Q$ and $W_V$ are attentional learning parameters. After ATS receives all key-value $(K^j, V^j)$ from all of teachers $j \in \mathcal{G}$, the attention weight $\alpha^{ij}$ is assigned by passing key vector from teacher and query vector from student into a softmax:
\begin{equation}
\alpha^{ij} = softmax \left(\frac{Q^{i}K^{j}}{\sqrt{D_K}}\right)
\end{equation}
Here, $D_K$ is the dimension of teacher $j$'s key vector, which is used to resolve vanishing gradients (Vaswani et al. 2017). The final policy advice is a  weight sum with a linear transformation:
\begin{equation}
v^i = W_T\sum_{j \neq i} \alpha^{ij} V^{j}
\end{equation}
Here, $W_T$ is a learning parameter for policy parameter decoding.

Behind the single attention head, we use a simple multi-attention head with a set of learning parameters $(W_K, W_Q, W_V)$ to aggregate all advice from different representation subplaces. Besides, attention head dropout is applied to improve the effectivity of our attention mechanism.

Finally, student agent $i$ obtains its action at this time with policy parameters from Attention Teacher Selector:
\begin{equation}
\widetilde{a}^i_t = v^i(m^i_t)
\end{equation}
In our experiments, the attention parameters $(W_K, W_Q, W_V)$ are shared across all agents, because knowledge transfer process is similar in all pairs of student-teacher, but different observations introduce different teacher weight vector. This setting encourages our approach to learn more efficient and make our model easy to be extended in different settings, such as larger number of agents or a different environment.

In this work, we consider scenarios where other agents' learning experience is useful to a student agent. Feeding student's observation information and teacher's learning experience into our attention mechanism helps to select action with other agents' behavioral policy for the student agent. This module is an end-to-end knowledge transfer method without any decentralized learning parameter sharing. 
\subsection{Self-learning Mode}
If agent $i$'s student actor chooses self-learning mode, the student actor sends $i$'s encoding hidden state $m^i_t$ to the actor network. In self-learning mode, agents learn as a common individual agent. Each agent's policy in self-learning mode is independently trained by DDPG \cite{lillicrap2015continuous} algorithm.

Agent $i$’s critic network is updated by TD error, $\mathcal{R}$ is agent $i$'s transition set:
\begin{equation}  
\begin{split}
\mathcal{L}\left(\theta^{Q}\right)=\mathbb{E}_{m_t, a_t, r_t, m_{t+1} \sim \mathcal{R}}\left[\left(y_t-Q\left(m_t, a_t | \theta^{Q}\right)\right)^{2}\right],\\
y_t=r_t+\left.\gamma Q\left(m_{t+1}, a^{\prime} | \theta^{Q^{\prime}}\right)\right|_{a^{\prime}=\mu^{\prime}\left(m_{t+1} | \theta^{\mu^{\prime}}\right)}
\end{split}
\end{equation}

The policy gradient of agent $i$’s actor network can be derived as:
\begin{equation}  
\nabla_{\theta^{\mu}} J=\mathbb{E}_{m_t, a_t \sim \mathcal{R}}\left[\nabla_{a} Q\left(m_t, a_t | \theta^{Q}\right)|_{a_t=\mu(m_t)} \nabla_{\theta^{\mu}} \mu\left(m_t | \theta^{\mu}\right)\right].
\end{equation}

In games with discrete action space, in self-learning mode, we refer to the modified discrete version of DDPG suggested by \cite{lowe2017multi} in agents' actor-critic networks. Agents in self-learning update its actor network use,
\begin{equation}\nabla_{\theta^{\mu}} J=\mathrm{E}_{m_t, a_t \sim \mathcal{R}}\left[\nabla_{a} Q(m_t, a_t) \nabla_{\theta^{\mu}} a_t\right]\end{equation}.

Our framework is adapted for both continuous action space and discrete action space.

\section{Empirical evaluations}
We construct three environments to test the team-wide performance of PAT and existing advising methods in multi-tasks and joint task scenarios. Also, we compare the scalability of all the approaches with the increase of number of agents. Additionally, the transferability of PAT is evaluated in different environments.

\subsection{Setup}
Empirical evaluations are performed on three cooperative multi-agent environments: Grid Treasure Collection, Moving Treasure Collection, Predator-Prey, and. We implement Grid Treasure Collection, a standard grid world environment. Moving Treasure Collection and Predator-Prey are implemented based on Multi-Agent Particle Environment \cite{mordatch2018emergence,lowe2017multi} where agents move around in a 2D space and involve interaction between agents. We briefly describe the three environments below:

\subsubsection{Grid Treasure Collection}
\begin{figure}
\centering
\includegraphics[width=0.30\textwidth]{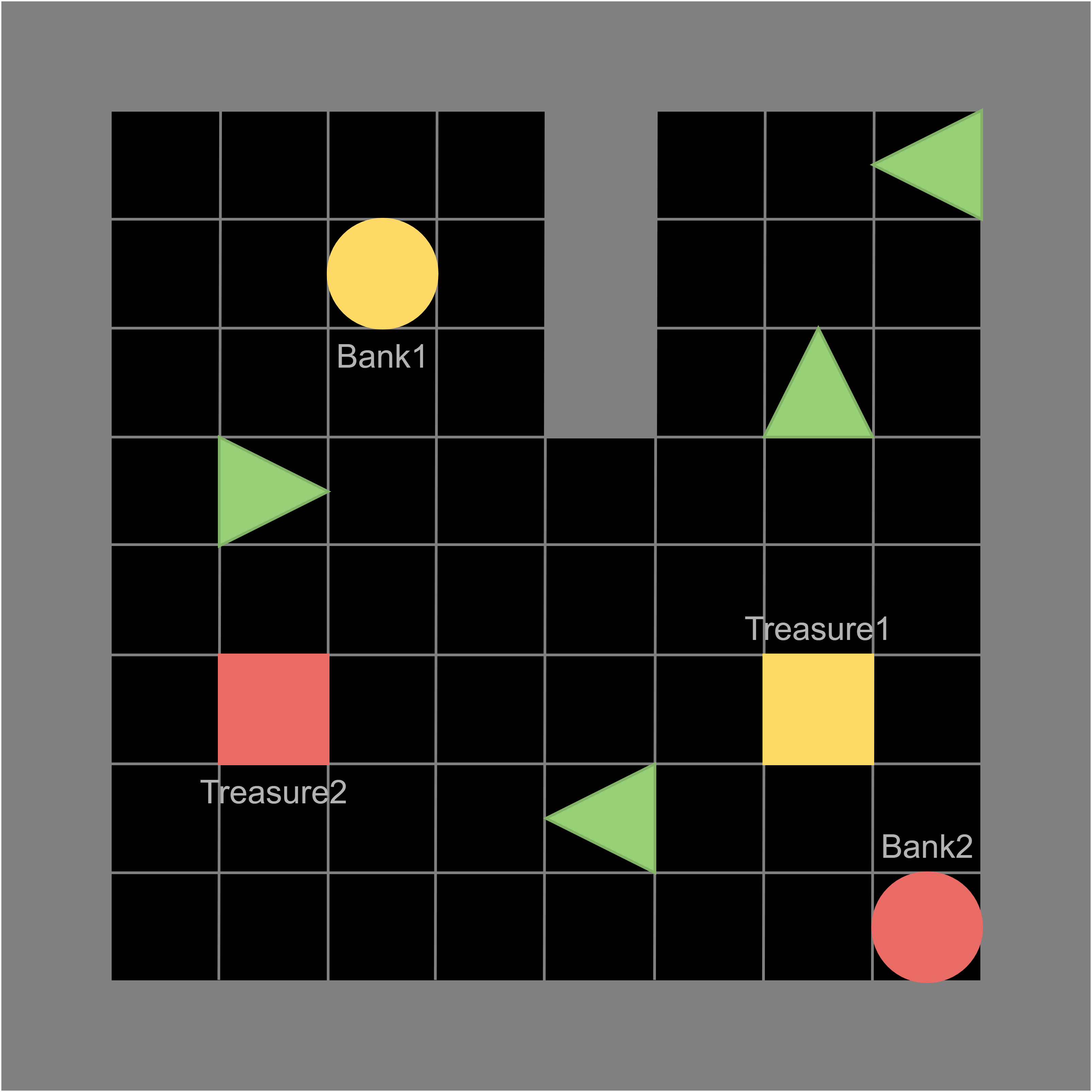}
\caption{Grid Treasure Collection, 4 agents, 2 treasure and 2 treasure banks}
\label{fig:Game Environment}
\end{figure}

: There are M agents and M/2 treasure grids and M/2 treasure banks in the grid maze (shown in Fig.~\ref{fig:Game Environment}). Each treasure is corresponding to a treasure bank. When agents collect treasures in treasure grids, agents get small rewards, and then return treasures to its corresponding bank, agents receives big rewards. Each treasure grid has M treasures and agents can only obtain one treasure from each treasure grid. The ability of agents to carry treasures is not limited. But when agents return wrong treasures to treasure banks, agents receive big penalties.

\subsubsection {Moving Treasure Collection}: The game rule in Moving Treasure Collection is similar to the above game, but in this environment, agents are green and treasures and treasure banks are moving randomly. And all objects are moving in a 2D open ground. Obstacles (large black circles) block the way in the environment.

\subsubsection{Cooperative Navigation}: There are M agents (green) and M landmarks (purple) in this environment. Agents are rewarded based on how far any agent is from each landmark. Agents are required to position themselves covering all the landmarks. When an agent covers a landmark, it gets a local reward. Obstacles (large black circles) block the way. 

To simplify our approach training, we set discrete action spaces in all the environments, allowing agents to move up, down, left, right, or stay. All agents receive partial observation at each step, and get local feedback from environments. We plan to evaluate our approach in both multi-task environment (Treasure Collection) and joint task environment 

\begin{figure}[t]
    \centering
    \includegraphics[width=0.50\textwidth]{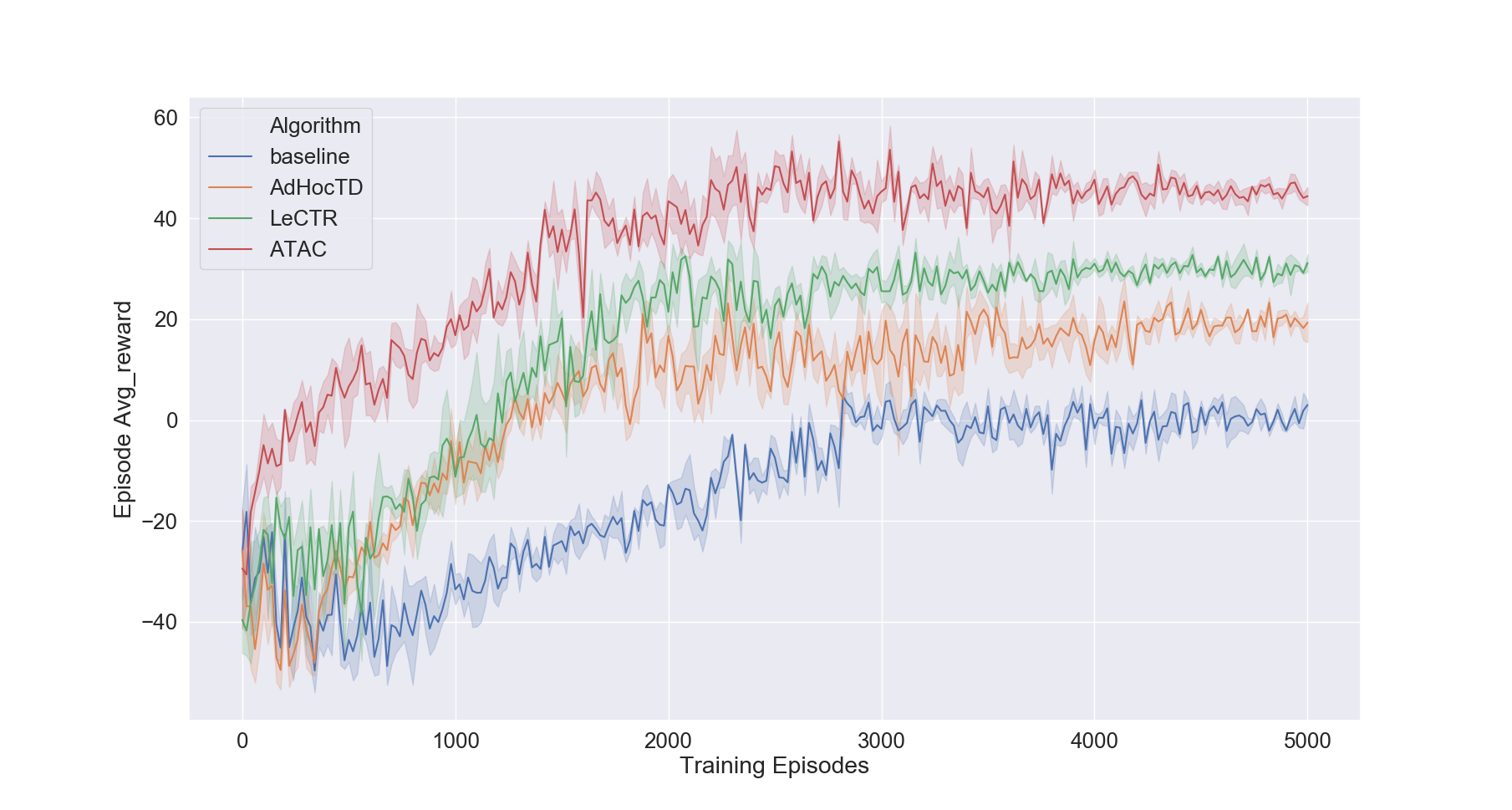}
    \caption{Average reward per episode in Grid Treasure Collection}
    \label{fig:Grid Collection}
\end{figure}

\subsection{Baselines}
We compare our approach, PAT with implementation of fully decentralized training approaches including Individual Deep Q-Learning (IQN) \cite{mnih2015human}, AdHocTD \cite{da2017simultaneously} and LeCTR \cite{omidshafiei2019learning}. IQN, as a distributed baseline,  is a reinforcement learning algorithm for single agent, which is trained independently for each agent with partial observation in our environments. AdHocTD, LeCTR and our method rely on action advising without any other information communication , which means each agent is unaware of the observations and rewards of other agents in the environment. We implement AdHocTD and LeCTR with original experimental parameters and public implementations. In LeCTR implementation, we use Majority Vote for multiple advice selecting. Hyperparameters are tuned based on our environments and performance. All implementation details will be released with code.

We evaluate all the models on two different agent teams , with M=4, M=8 and M=12 agents. When the number of agents increasing, the amount of calculation and difficulty increase rapidly.. Average step length per episode (Treasure Collection games' max episode length = 1000), success rate of covering all landmarks in Cooperative Navigation, and average sum of rewards per episode are three indicators used to evaluate performance of all models. All results are reported using 30 independent training setting different seeds. Table I shows the results in 4-agent and 12-agent environments for comparison. (The results in 8-agent environment will be presented in later released data)
\begin{table*}[t]
\centering
\renewcommand\arraystretch{1.3}
\setlength{\tabcolsep}{0.20cm}{
\caption{Evaluation Results}
\begin{tabular}{cccccccc}
\hline
TASK                       & APPROACH & \begin{tabular}[c]{@{}c@{}}M=4\\ Avg\_step\end{tabular} & Success \%   & Avg\_reward    & \begin{tabular}[c]{@{}c@{}}M=12\\ Avg\_step\end{tabular} & Success \%   & Avg\_reward    \\ \hline
Grid Treasure Collection   & Baseline & $1000\pm 0$                                             & -            & $0.78\pm1.43$  & $2000\pm0$                                              & -            & $-1.20\pm0.04$ \\
Grid Treasure Collection   & AdHocTD  & $876 \pm 15$                                            & -            & $18.76\pm1.14$ & $1821\pm29$                                             & -            & $12.56\pm0.34$ \\
Grid Treasure Collection   & LeCTR    & $812 \pm 39$                                            & -            & $29.87\pm1.23$ & $1836\pm23$                                             & -            & $11.68\pm2.24$ \\
Grid Treasure Collection   & PAT     & $762 \pm 20$                                            & -            & $45.65\pm2.32$ & $1157\pm37$                                             & -            & $20.34\pm1.49$ \\ \hline
Moving Treasure Collection & Baseline & $1000 \pm 0$                                            & -            & $-1.84\pm0.66$ & $2000\pm0$                                              & -            & $-3.59\pm0.26$ \\
Moving Treasure Collection & AdHocTD  & $1000 \pm 0$                                            & -            & $8.22\pm1.79$  & $1987\pm67$                                             & -            & $-3.34\pm0.21$ \\
Moving Treasure Collection & LeCTR    & $877 \pm 46$                                            & -            & $20.76\pm0.24$ & $1889\pm48$                                             & -            & $-2.79\pm2.12$  \\
Moving Treasure Collection & PAT     & $854\pm 23$                                             & -            & $33.98\pm2.13$ & $1239\pm32$                                             & -            & $-0.32\pm0.43$  \\ \hline
Cooperative Navigation     & Baseline & $397 \pm 25$                                            & $52.2\pm2.0$ & $-1.78\pm0.03$ & $782\pm4$                                               & $32.7\pm4.7$ & $-4.68\pm0.05$ \\
Cooperative Navigation     & AdHocTD  & $320 \pm 28$                                            & $69.0\pm3.7$ & $-1.23\pm0.27$ & $547\pm29$                                              & $42.6\pm2.0$ & $-3.50\pm0.08$ \\
Cooperative Navigation     & LeCTR    & $278 \pm 25$                                            & $81.4\pm3.2$ & $-1.29\pm0.39$ & $672\pm14$                                              & $40.7\pm1.9$ & $-3.36\pm0.70$ \\
Cooperative Navigation     & PAT     & $289 \pm 18$                                            & $80.2\pm3.2$ & $-0.45\pm0.07$ & $498\pm10$                                              & $59.2\pm1.3$ & $-2.67\pm0.02$ \\ 
\hline
\end{tabular}}
\end{table*}

\subsection{Results and Analysis}
Fig.~\ref{fig:Grid Collection} displays average team rewards per episode in Grid Treasure Collection. A thorough evaluation result is summarized in Table I. 

In 4-agent Grid Treasure Collection (GTC) and Moving Treasure Collection (MTC)  game, PAT outperforms all models with a higher average reward after convergence. In Comparison of all approachs in average episode length (Ave\_step), PAT has no obvious advantage compared with LeCTR, but PAT greatly improves average team reward per episode (Avg\_reward). Attention mechanism performs well and significantly fights out agents with useful experience (have successfully collected treasures), which can correctly guide unfamiliar agents to take more beneficial actions. AdHocTD selects teachers using the number of times the agent visited the current state. As a result, the phenomenon of over-advising appears in AdHocTD after trained over 2000 episodes, which means that students may take advice from teachers with more bad experience. It contributes to AdHocTD worse final reward performance than LeCTR and PAT. LeCTR  learns how to teach by centralized training and decentralized executing, causing the learning instability in early episodes. Observing the results, PAT successfully avoids this problem by training a decentralized student actor-critic network for the decision of acting mode.  

In 4-agent Cooperative Navigation, PAT surprisingly performs longer episode length and lower success rate than LeCTR. Attention mechanism in PAT tends to imitate successful action from other agents, which helps student agent gain more local rewards. But it might cause a bad impact on team coordination in a joint task. For example, agents tend to cover the same landmarks with successful experience, but the team needs to cover all the landmarks. PAT is hard to learn the team cooperative policy and gain more cooperative rewards. However, LeCTR uses a centralized critic network to calculate advising value concatenating all agents' observation, which exactly improves cooperation performance between agents. This result gives us a future direction that we should try to modify attention tendency in global reward for cooperative tasks.

\subsubsection{Scalability}

When the number of agents is 12 in all the environments, because of the increasing computation difficulty in selecting advice from a larger agent team, AdHocTD's advising probabilistic mechanism can not handle the problem in more information selecting, resulting in sub-optimal rewards. LeCTR, as an algorithm originally supporting two-player games, has low performance in a larger size of agent team. We suspect it due to LeCTR's centralized control of advising. With more agents in the environments, LeCTR's centralized advising-level critic performs poorly with limited training time. As expected,  PAT's attention selector effectively filtering knowledge from more teachers and maintain student mode's accuracy, so our approach has a larger advantage with the increase in number of agents. Our experimental results confirm our inference.

In summary, PAT performs much better than all approaches and has a distinct advantage in average team-wide rewards in complex multi-task environments as expected. We also report PAT's disadvantage in the joint-task scenario. PAT scales better when agents are added in all the experiments, which shows that sharing attention mechanism is useful for information selecting in a large multi-agent team.

\subsubsection{Transferablility}
Analysing the above results, the behavior knowledge transfer becomes more difficult when number of agents increases. We design a new experiment to explore our model parameters transfer performance. We test our approach in two experiments with different numbers of agents. First, agents are trained in an environment with a small number of agents. Then, the trained parameters of agents' shared attention mechanisms are transferred/reused in an environment with more agents. We compared our transfer experimental performance with original training performance data of larger agent teams.

Table II shows the transfer performance of 4-agent attention mechanism in 8-agent environment, and Table III presents the comparison between original learning performance and 6-agent model transfer data in 12-agent environment. In transfer experiments, the team of 8 or 12 agents trains their new individual student actor-critic network and self-learning network based on the transferred attention mechanism trained by a smaller team.

According to the compared results summarized in Table II and III, our approach can be efficiently transferred. The transferred model successfully achieve nearly 90\% of the original training performance, which also better than all other approaches with fully training. It can save large computational costs for large-size team of agents.
. Our shared attention selector can effectively solve new tasks based on related experience.

\begin{table}[!htbp]
\caption{Transfer Evaluations in M=8}
\label{tab:Transfer}
\begin{tabular}{ccccc}
\hline
\begin{tabular}[c]{@{}c@{}}Task\\ M=8\end{tabular} & \begin{tabular}[c]{@{}c@{}}Original\\ Avg\_step\end{tabular} & Avg\_reward    & \begin{tabular}[c]{@{}c@{}}M=4\\ Transfer\\ Avg\_step\end{tabular} & Avg\_reward    \\ \hline
Grid                                               & $914\pm65$                                                   & $27.80\pm3.49$ & $1021\pm18$                                                         & $24.86\pm2.45$ \\
Moving                                             & $1174\pm42$                                                  & $10.79\pm3.58$ & $997\pm10$                                                         & $9.37\pm3.02$  \\ \hline
\end{tabular}
\end{table}

\begin{table}[!htbp]
\caption{Transfer Evaluations in M=12}
\label{tab:Transfer}
\begin{tabular}{ccccc}
\hline
\begin{tabular}[c]{@{}c@{}}Task\\ M=12\end{tabular} & \begin{tabular}[c]{@{}c@{}}Original\\ Avg\_step\end{tabular} & Avg\_reward    & \begin{tabular}[c]{@{}c@{}}M=8\\ Transfer\\ Avg\_step\end{tabular} & Avg\_reward    \\ \hline
Grid                                                & $1157\pm37$                                                  & $20.34\pm1.49$ & $1230\pm9$                                                         & $18.85\pm0.98$ \\
Moving                                              & $1239\pm32$                                                  & $-0.32\pm0.43$ & $1389\pm24$                                                        & $3.06\pm0.09$  \\ \hline
\end{tabular}
\end{table}

\section{Conclusions and Future Work}
We introduce a parallel knowledge-transfer framework, PAT for decentralized multi-agent reinforcement learning. Our key idea is designing two acting mode for agents and using a shared attention mechanism to select behavior knowledge from other agents to accelerate student agent learning. We empirically evaluate our proposed approach against all state-of-the-art advising or teaching methods in multi-agent environments. Results in experiments of scaling the number of agents and model transfer are also shown. Extending knowledge transfer in joint task learning and more complicated multi-agent systems is our future research direction.


\bibliographystyle{IEEEtran}
\bibliography{reference.bbl}
\end{document}